\documentclass[10pt,twocolumn,letterpaper]{article}
\usepackage{wacv}
\usepackage{times}

\usepackage{graphicx}
\usepackage{comment}
\usepackage{amsmath,amssymb} 
\usepackage{color}

\usepackage{multirow}
\usepackage{booktabs}

\usepackage{epsfig}
\usepackage{adjustbox}
\usepackage{subfigure}
\usepackage{bm}

\wacvfinalcopy 

\ifwacvfinal
\def\assignedStartPage{1} 
\fi

\ifwacvfinal
\usepackage[breaklinks=true,bookmarks=false]{hyperref}
\else
\usepackage[pagebackref=true,breaklinks=true,colorlinks,bookmarks=false]{hyperref}
\fi

\ifwacvfinal
\else
\fi

\begin{document}

\title{Attention-Aware Noisy Label Learning for Image Classification} 

\author{
Zhenzhen Wang\\
Nanyang Technological University \\
Singapore\\
{\tt\small zwang033@e.ntu.edu.sg}
\and
Chunyan Xu\\
Nanjing University of Science and Technology\\
Nanjing, China\\
{\tt\small cyx@njust.edu.cn}
\and
Yap-Peng Tan\\
Nanyang Technological University\\
Singapore\\
{\tt\small eyptan@ntu.edu.sg}
\and
Junsong Yuan\\
State University of New York at Buffalo\\
NY, USA\\
{\tt\small jsyuan@buffalo.edu}
}
\maketitle

\begin{abstract} 
Deep convolutional neural networks (CNNs) learned on large-scale labeled samples have achieved remarkable progress in computer vision, such as image/video classification. The cheapest way to obtain a large body of labeled visual data is to crawl from websites with user-supplied labels, such as Flickr. However, these samples often tend to contain incorrect labels (i.e. noisy labels), which will significantly degrade the network performance. In this paper, the attention-aware noisy label learning approach ($A^2NL$) is proposed to improve the discriminative capability of the network trained on datasets with potential label noise. Specifically, a Noise-Attention model, which contains multiple noise-specific units, is designed to better capture noisy information. Each unit is expected to learn a specific noisy distribution for a subset of images so that different disturbances are more precisely modeled. Furthermore, a recursive learning process is introduced to strengthen the learning ability of the attention network by taking advantage of the learned high-level knowledge. To fully evaluate the proposed method, we conduct experiments from two aspects: manually flipped label noise on large-scale image classification datasets, including CIFAR-10, SVHN; and real-world label noise on an online crawled clothing dataset with multiple attributes. The superior results over state-of-the-art methods validate the effectiveness of our proposed approach.  
\end{abstract}
	
\section{Introduction}
	
\begin{figure}
	\centering
	\includegraphics[width=0.8\linewidth]{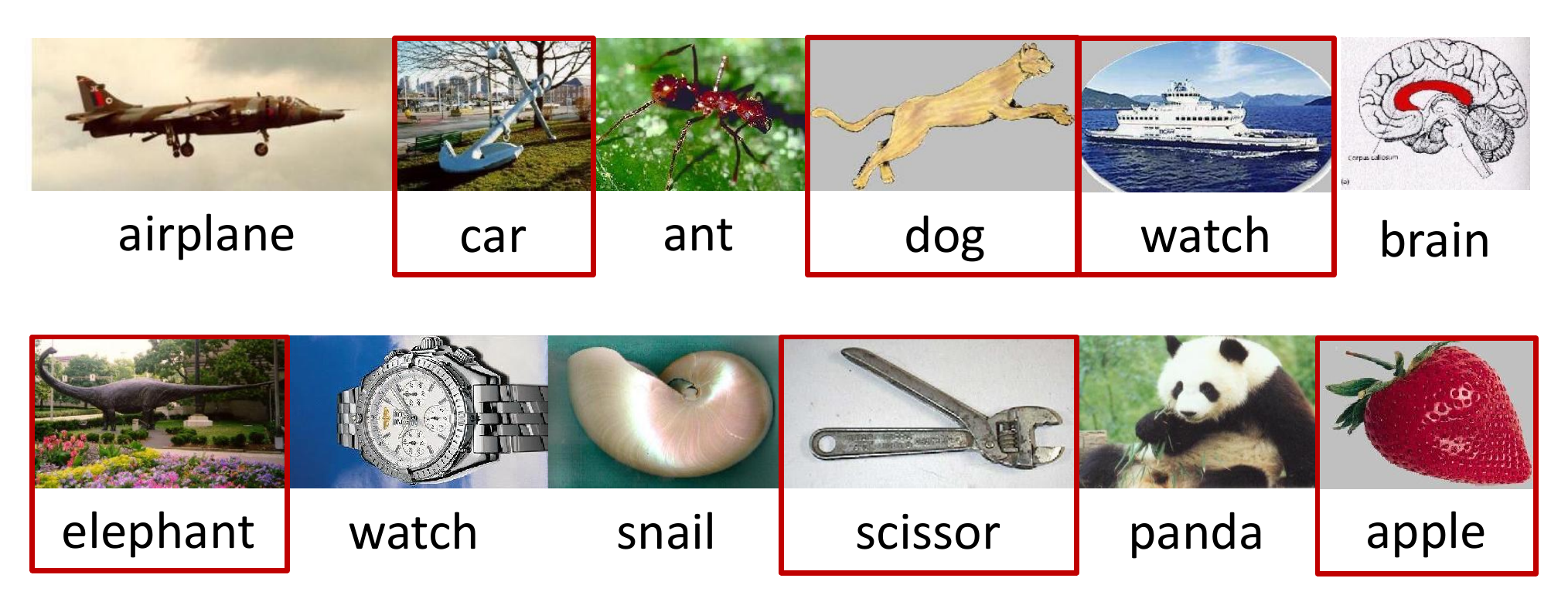}
	\caption{Examples of noisy labels. The image annotations that are manually labeled by amateurs or automatically generated by a machine are not reliable. Worse, the noisy labels and the number of mislabeled images are not specified.}\label{fig:noisylabels}
	
\end{figure}

CNNs have triumphed over many vision tasks. However, the overwhelming performances of CNNs heavily rely on large-scale high-quality labeled data, e.g., ImageNet \cite{ILSVRC15}, which are typically laborious and costly to collect and annotate. Nevertheless, there are millions of freely available images with user-supplied labels that can be easily accessed from the web. Although utilizing web images has become a popular research direction in the field of large scale image recognition, the performance is obviously inferior to its counterpart on finely labeled data. Directly using image sets with a high proportion of noisy labels (e.g., Fig.~\ref{fig:noisylabels}) can even degrade the performance of finely-trained CNN models~\cite{gong2013deep,Xiao_2015_CVPR,yao2017exploiting}. Thus, it is highly desired to design a network that is able to mitigate the impact of noisy labels.

There are already several label noise-robust algorithms being developed in recent years. Some researchers propose robust loss functions specifically for noisy image classification \cite{liu2017self,tanaka2018joint}, others try to improve the quality of training data by predicting the label noise type or removing the mislabeled samples \cite{Xiao_2015_CVPR,wang2018iterative}. However, these methods either work worse under large proportions of label noise or require prior knowledge on the patterns of label noise. 
There are also CNN-based methods explicitly modeling the noisy distributions by a noise layer \cite{sukhbaatar2014learning,jindal2016learning,goldberger2017training}. However, these methods are usually based on the assumption that the disturbance of all samples in the same class is equal, thus they are unable to acquire diverse noisy information, e.g., some furry dogs are easy to be labeled as cats and some large ones as horses, which are very common for online images with user-supplied labels.  

In this work, we propose an attention-aware noisy label learning approach, termed $A^2NL$, to improve the discriminative capability of noise-robust network. In particular, a Noise-Attention (NA) model is proposed to explore multiple distributions of label noise and a recursive learning strategy is employed to further boost its learning ability. Both contributions can be applied to any conventional CNNs. To avoid confusion, the network with the NA model is informally called attention network in the following. Unlike previous works which describe the noisy-label information of an image set by one distribution, our NA model contains multiple units, each of which pays attention to a specific noisy-label distribution. The reasons for such an improvement are two folds. 1) The noisy levels of different classes vary a lot, e.g., in the task of predicting clothes color, it is likely to mix the orange and brown while it is almost impossible to label a red one as a blue one. 2) The individual images of the same class can even present different disturbance. For example, some dogs could be clearly recognized, some may be labeled as cats or horses. By modeling multiple noisy-label distributions, the proposed NA model can not only portray the different noisy levels among classes, but also distinguish the diverse noisy-label distributions among images.

The proposed recursive learning strategy is inspired by \cite{xu2018srnn,hinton2015distilling,bucilu2006model} that the soft predictions of a well-trained classifier usually contain rich information. The soft outputs not only indicate the object class of the input image, but also reflect the relations among classes. For example, if a cloth sample is predicted as an orange one with the confidence of 80\%, a brown one with 15\%, and 5\% for other classes. The biggest figure (80\%) advocates that the input image contains a cloth in orange, and other figures suggest that the orange is highly possible to be mixed up with the brown. Thus, to boost the learning ability of the proposed attention network, we recursively train it by distilling the knowledge from a well-trained attention network in the previous iteration. To be specific, the outputs of the attention network in the previous iteration, coupling with the given training labels, are combined as the training supervisions for the network in the current iteration. Different from directly combining multiple network models, the recursive learning strategy is able to assemble the network knowledge in previous iterations without introducing more parameters.     

We conduct extensive experiments on both datasets with synthesized noisy labels (randomly flipping the labels) including SVHN \cite{netzer2011reading}, CIFAR-10 \cite{krizhevsky2010convolutional}, and a real-world clothes dataset with multiple attributes \cite{huang2015cross} which naturally contains mislabeled samples. As considering both general and specific label noise simultaneously, the proposed framework shows excellent effectiveness and robustness to both synthesized and real-world label noise. Our main contributions are summarized as follows:

\begin{itemize}
	\item A Noise-Attention model with multiple noise-specific units is proposed to explore various distributions of label noise, which can be applied to conventional CNNs. 
	\vspace{-2mm}
	\item A recursive learning strategy, which could assemble the high-level knowledge learned from multiple networks, is introduced to boost the attention network learning ability.\vspace{-2mm}
	\item Extensive experiments on both manually flipped label noise and real-world label noise demonstrate the excellent effectiveness and robustness of our attention-aware noisy label learning framework.

\end{itemize}

\section{Related Works}
\label{sect:related_work} 

Label noise is common, it refers to the associated labels of instances might be incorrect, e.g., a truck is labeled as train, which is inevitable due to imperfect evidence, patterns or insufficient information for reliable labeling. Compared with weak-/semi-supervised learning problem where high-quality samples are specified, the noisy-label classification is more challenging as we do not know which samples are incorrectly labeled, neither the total number of mislabeled samples.

\begin{figure*}
	\centering
	\includegraphics[width=0.8\linewidth]{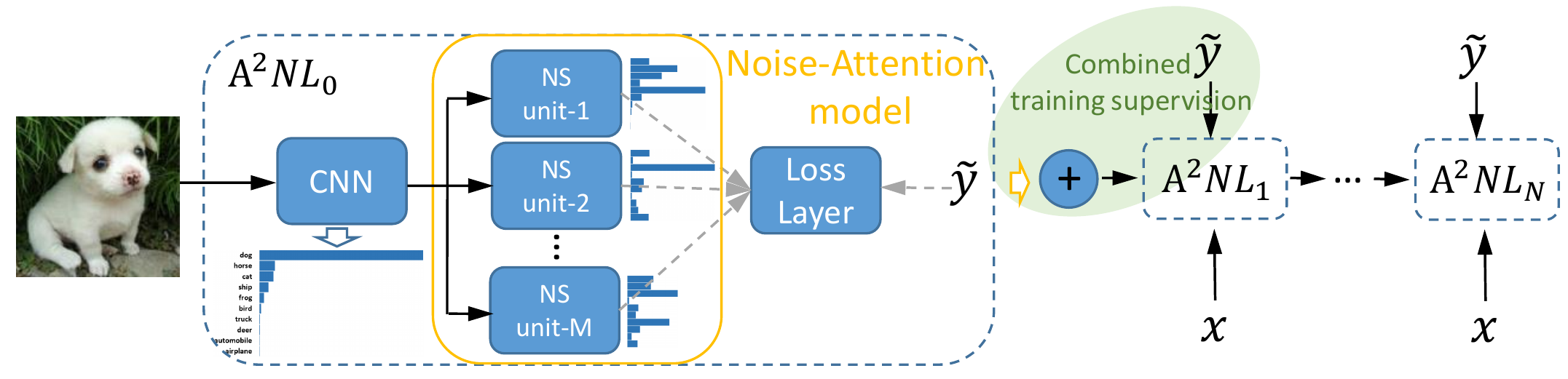}   
	\caption{An overview of the proposed attention-aware noisy learning framework. First, a Noise-Attention model is introduced, which consists of multiple noise-specific units (NS unit) with each paying attention to a specific noisy distribution. The attention network is trained by the proposed recursive learning strategy. The training supervision is updated at each recursive iteration, it is combined by the given labels and the outputs from previous iteration.} 
	\label{fig:multi-noise}
	
\end{figure*}    

\textbf{Noisy Samples Cleaning:} The most straightforward way is to improve the quality of image collections by label noise cleansing skills. For example, Jeatrakul \textit{et al.}~\cite{jeatrakul2010data} learn a Neural Network using training data and remove all instances that are misclassified by the Network. Wang \textit{et al.} \cite{wang2018iterative} use a Siamese network to learn discriminative features from predicted clean and noisy instances. Alternatively, the learned features will be used to train a noise detector. However, those methods suffer from a contradictory dilemma, i.e., the datasets with label noise often lead to poor classifiers, whereas high-quality classifiers are necessary for precise predictions. Thus removing or relabeling mislabeled instances may accumulate noises.

\textbf{Robust Loss Designing:} To build models that are intuitively robust to label noise is another solution. Bootkrajang and Kaban~\cite{bootkrajang2013boosting} derive two noise robust approaches from AdaBoost, one is to construct a robust classifier through a probabilistic latent variable model, the other is to modify the AdaBoost algorithm by combining two exponential losses as a new objective function. Semi-supervised learning~\cite{weston2012deep} and Zero shot learning~\cite{jayaraman2014zero} are also popular algorithms to address label noise. Pan \textit{et al.} \cite{pang2016robust} propose the robust latent poisson deconvolution for topical detection from noisy web images. However, the main problem of the aforementioned approaches is that they increase the complexity of learning algorithms and can lead to overfitting.                   

\textbf{Noisy Distributions Modeling:} To explicitly model the noisy distributions of given image sets is another approach. Based on the assumption that all instances from a dataset following the same noisy distribution, Sukhbaatar \textit{et al.}~\cite{sukhbaatar2014learning} model the noisy distribution by a noisy layer, which is set on top of the softmax layer. Inspired by them, Xiao \textit{et al.} \cite{Xiao_2015_CVPR} build a probabilistic graphical model to learn the relationships among images, class labels and label noise. Jacob and Ehud \cite{goldberger2017training} propose similar solution but optimize the CNN with EM algorithm rather than the conventional SGD. Our proposed method falls into this category with goals to learn the noisy distributions. But different from previous methods, we propose a Noise-Attention model implementing with multiple noise-specific units to cater to different noisy distributions. In addition, a recursive learning strategy is designed to boost the learning ability of the proposed noise-attention network.

\section{Attention-Aware Noise Learning}
\label{sect:method}

The goal of this work is to learn a classifier based on the given noisy data. To this end, we introduce an attention-aware noise learning framework ($A^2NL$) which consists of a Noise-Attention model to learn noisy distributions and a recursive learning strategy to boost the discriminative capability of the learned network (see Fig.~\ref{fig:multi-noise}).   

\subsection{Noise-Attention model}
\label{sect:noisemodel}

A typical classification network is usually followed by a softmax layer at the end, the outputs of which indicate the class confidences predicted by the network. For an input sample $\{x_n, y_n\}$, the output of softmax layer is written as: $\bm{p}=\{p(y=j|x,\theta)\}_{j=1}^C$, where $C$ is the number of classes, $\theta$ represents the CNN parameters. The learning objective is to minimize the negative log likelihood over $N$ samples:  

\begin{equation}\label{equat:softmaxloss}
L(\theta)=-\frac{1}{N}\sum_{n=1}^{N}\log p(y=y_n|x_n,\theta)
\end{equation}  

\begin{figure}
	\centering
	\includegraphics[width=0.6\linewidth]{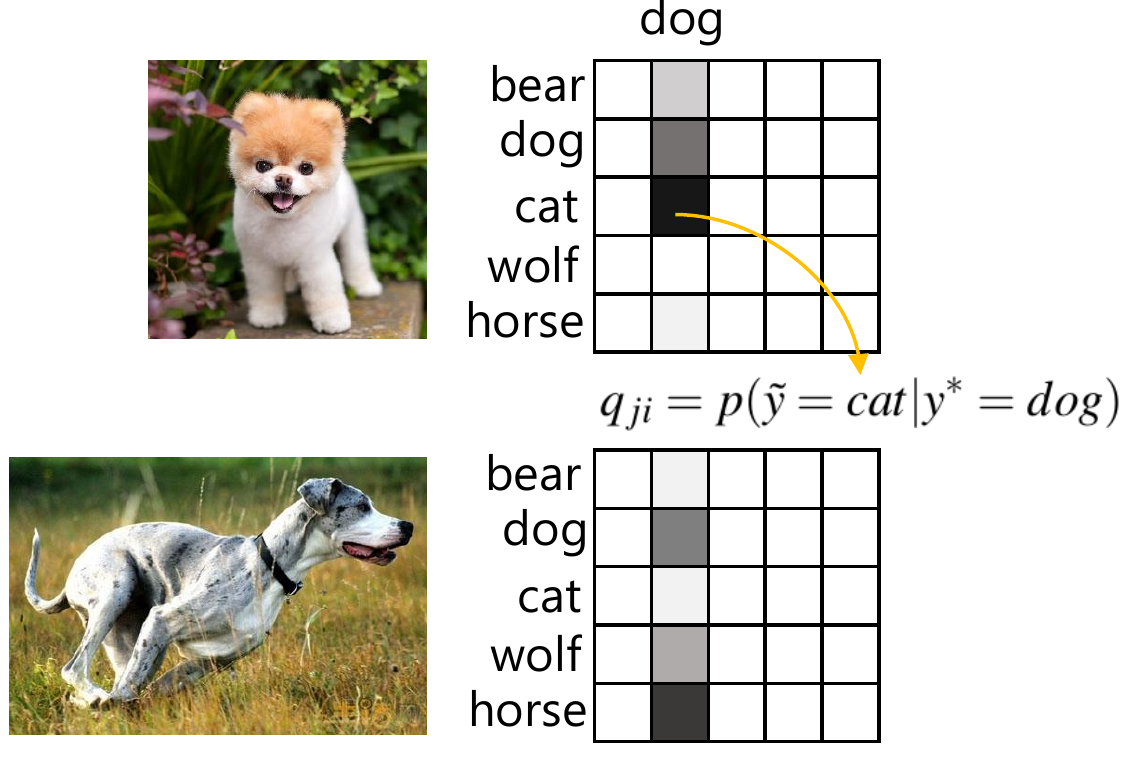}
	
	\caption{Illustration of the noise-specific unit.}\label{fig:qji}
	
\end{figure}

Different from traditional object recognition tasks, we consider in this paper a more challenging situation in which the image sets may contain noisy labels. Let $\{(x_n,\tilde{y}_n)\}_{n=1}^{N}\subset\tilde{D}$ denote the noisy dataset, where $x_n$ is the $n$-th training sample and $\tilde{y}_n\in\{1,2,...,C\}$ is the given label (may be mislabeled). The true label is denoted by $y^*_n\in\{1,2,...,C\}$, which is treated as a latent variable. Then the conditional distribution of the given label on the true label of a sample is described by $Q=\{q_{ji}\}_{i,j=1}^{C}, q_{ji}:=p(\tilde{y}=j|y^*=i)$, where $Q$ is a $C\times C$ matrix with the sum of each column equal to 1. Different from~\cite{sukhbaatar2014learning,goldberger2017training}, which describe the noise in the whole dataset with only one noisy distribution, the proposed NA model contains multiple noisy units with each paying attention to a specific noisy distribution, parameterized by matrix $Q_m$:
\begin{equation}\label{equat:multi-Q}
\begin{cases}
Q_m =I & \text{ if } m=1\\
Q_m \neq I &\text{ if } m>1
\end{cases}
\end{equation}
where $I$ is an identity matrix, $Q_m$ is the noise distribution matrix in the $m$-th unit, $m=\{1,2,...,M\}$, $M$ is the total number of noise-specific units. As shown in Fig.~\ref{fig:qji}, the furry dog is easy to be labeled as a cat, and the big one tends to be mislabeled as a horse, thus multiple label noise distributions are necessary for catering different prototypes. Here, $Q_1=I$ is fixed to represent the noisy distribution of clean samples, while other noise-specific units will update during training to learn the noisy distributions. 

In order to apply the proposed NA model to CNNs, we implement the noisy units as fully-connected layer without bias, the weight of each unit, $Q_m$, denotes a confusion matrix between observed labels and latent true labels. Fig.~\ref{fig:multi-noise} highlights the proposed NA model. For an input sample $\{x_n, \tilde{y}_n\}$, it first passes through a conventional CNN model (called base model), and the proposed NA model sequentially. The intermediate outputs from base CNN pass through all the $M$ noisy units and thus generate $M$ outputs: $\{\bm{p}_m\}_{m=1}^M$, $\bm{p}_m\in\mathbb{R}^{C\times1}$. Instead of using all noisy units in the NA model, we only use the one that maximizes the prediction confidence, i.e., $Q_{max}=\arg\max_Q p(y=\tilde{y}_n|x_n,\theta,Q)$. Therefore, the confidence of an input $x_n$ being predicted as class $j$ by the attention network is given by:  
\begin{equation}
p(y=j|x_n, \theta, Q_{max}),
\end{equation}
then the negative log likelihood loss for the attention network can be written as:
\begin{equation}\label{equat:costfun01}
L(\theta,Q)=-\frac{1}{N}\sum_{n=1}^{N}\log p(y=j|x_n, \theta, Q_{max})
\end{equation}  

We employ stochastic gradient descent~(SGD) with back-propagation technique to optimize the parameters in the attention network. Since the NA model consists of linear layers, the gradients can pass through them to the base network. Before learning the NA model, we first train the base network with noisy data to its optimal so that the noisy information could be well absorbed. Then the noise-specific units parameterized by  $\{Q_m\}_{m=1}^M$ are added gradually to the base model, with each being initialized as identity (matrix), $Q_m=I$. A new noise-specific unit will be involved and updated with larger decay once the performance improvement is limited with the current unit. By this training strategy, the noisy information absorbed by the base network in the first dozens of epochs is expected to gradually distill to the NA model. In addition, the noise-specific units are forced to learn diverse noisy distributions with sequentially adding strategy and different learning decay values, which are experimentally verified in Sect. \ref{sect:experiment}. 

The goal of the proposed method does not end with predicting the noisy labels $\tilde{y}$ or learning the noisy distributions, instead it is to predict the true labels given the image sets with noisy labels. Recall that the true label $y^*$ is regarded as a latent variable, its relation with the noisy labels and noisy distributions can be expressed as:
\begin{equation}\label{equat:Qprediction}
\begin{aligned}
&p(\tilde{y}=j|x_n,\theta,Q_{max})\\
=&\sum_{i=1}^{C}p(\tilde{y}=j|y^*=i)p(y^*=i|x_n,\theta)\\
=&\sum_{i=1}^{C}q_{ji}p(y^*=i|x_n,\theta)
\end{aligned}
\end{equation}
where $p(\tilde{y}=j|x,\theta,Q_{max})$ and $Q_{max}=\{q_{ji}\}_{i,j=1}^C$ denote the predictions of the attention network and the noisy distribution. Obviously, $p(y^*=i|x_n,\theta)$ is the prediction of base network. Therefore, if the given training labels are correctly predicted (i.e., $p(\tilde{y}=\tilde{y}_n|x_n,\theta,Q_{max})=1$) and the noisy distributions are exactly learned (i.e., $q_{\tilde{y}_n,y_n^*}=1$, $q_{j\neq\tilde{y}_n,i\neq y_n^*}=0$), then theoretically the base network is able to predict the true labels accurately \cite{sukhbaatar2014learning}. Although the NA model will introduce more parameters in training phase, it is similar to traditional classification network in inference with the outputs of the base network being used to evaluate.

\subsection{Recursive Learning}
\label{sect:seqmodel}   

In this part, we put forward a recursive learning algorithm to further boost the learning ability of the attention network. It is motivated by model compression strategies \cite{xu2018srnn,hinton2015distilling,bucilu2006model}, in which to compress an ensemble of models into a single model without significant loss in performance is possible by mimicking the function learned by the ensemble. The key idea of our proposed method is to fine-tune the attention network by leveraging the high-level knowledge from a well-trained network, so that it can make predictions similar to the ensemble of the two. In addition, the proposed recursive learning is in an iterative manner, with shared network architectures and dynamically updating training supervisions in each iteration, making the final network a compressed model of all previous networks. Specifically, the training supervisions in current iteration consist of two parts, 1) the fixed given training labels, and 2) the outputs of the attention network from the previous iteration. Here, we use ``iteration'' to denote one round in the recursive learning rather than the running iteration known in training CNNs. The network is initialized with parameters from the previous iteration, and is able to bootstrap itself with the updated training supervisions. The process continues until the improvement over the previous iteration is limited, the model obtained in the last iteration is used for inference. Fig.~\ref{fig:multi-noise} outlines our proposed recursive learning procedure. Compared with combining the models in all iterations for inference, our recursive learning strategy is smaller in size to store and faster in runtime to evaluate.     

\textbf{Sample-Wise Summation.} Inspired by~\cite{hinton2015distilling}, the network trained by the combined training supervisions can achieve a similar performance to an ensemble model combining two networks. We thus integrate the high-level knowledge in a well-trained attention network with the given labels as the new training supervisions. The combination is applied to guide the attention network training in the current iteration. For a given sample $\{x_n, \tilde{y}_n\}$, $p(y=j|x_n,\theta^{t-1},Q_{max}^{t-1})$ represents the confidence of being classified as class $j$ by the network from the $(t-1)$-th iteration. Let $\bm{s}_n=\{s_{n,j}\}_{j=1}^{C}$ denote the combined training supervisions of sample $x_n$, then the weighted combination is formulated as:
\begin{equation}\label{equat:sum}
s_{n,j}^t=\alpha^t\cdot\textbf{1}_{n,j} + p(y=j|x_n,\theta^{t-1},Q_{max}^{t-1})
\end{equation}
where $\textbf{1}_{n}$ is a binary vector with only one non-zero element indicating the given label of the input $x_n$, the subscript $j$ denoting its $j$-th element. Thus, the above formulation can be rewritten as:
\begin{equation}
\begin{aligned}
&s^t_{n,\tilde{y}_n}=\alpha^t+p(y=\tilde{y}_n|x_n,\theta^{t-1},Q_{max}^{t-1})\\
&s^t_{n,j\neq\tilde{y}_n}=p(y=j|x_n,\theta^{t-1},Q_{max}^{t-1})
\end{aligned}
\end{equation}
$\bm{s}_n\in\mathbb{R}^{C\times1}$ is then normalized with its elements summing to $1$. Also, only the noisy unit that maximizes the confidence of correctly classifying, $Q_{max}$, is used. The coefficient $\alpha$ is manually fixed with decreasing value as the network is expected to be more reliable with more iterations. Compared with directly combining two networks, such a combination could also integrate the knowledge in pretrained network while without introducing much more parameters.     

At the first iteration (denoted by $A^2NL$ in Fig.~\ref{fig:multi-noise}), the attention network is trained with given labels as in conventional classification task, all parameters of the network are trained from scratch. From the second iteration, the networks are initialized by parameters learned in the previous iteration, the training supervisions are composed of the given labels and the outputs of previously trained network. Fig.~\ref{fig:multi-noise} illustrates the recursive learning process. The network at the first iteration is trained with objective function presented in Eq.~\eqref{equat:costfun01}. From the second iteration, the objective function is to minimize the negative log likelihood with the updated training supervisions:
\begin{equation}\label{equat:costfun02}
L(\theta,Q)=-\frac{1}{N}\sum_{n=1}^{N}\sum_{j=1}^{C}s_{n,j}^t\log p(y=j|x_n,\theta^t,Q_{max}^t)
\end{equation}  

Although the proposed recursive learning method consumes more computing resources in the training phase, during inference the complexity does not increase as only the base network at the last iteration is used. Since in each recursive iteration the predictions from the previous iteration are integrated as supervisions, the learning ability will be distilled gradually, which yields the last network that makes predictions similar to the ensemble of models in all iterations \cite{bucilu2006model}. Thus, compared with existing works \cite{Xiao_2015_CVPR,wang2018iterative}, which require two independent networks in both training and inference, our proposed method is more efficient and effective during inference.     

\begin{figure}
	\centering
	
	\includegraphics[width=0.7\linewidth]{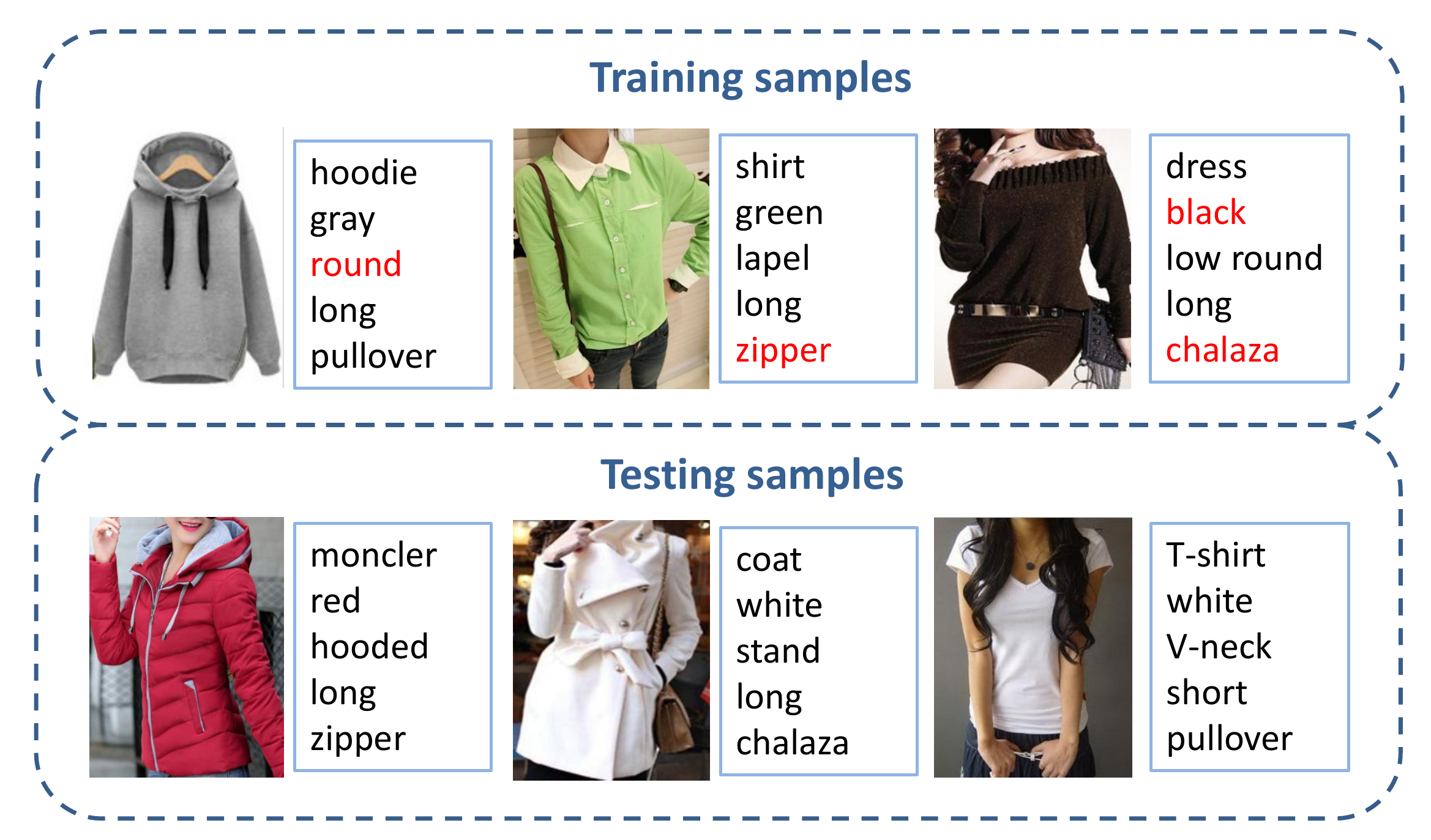}
	
	\caption{Cloth samples with multiple attributes.}\label{fig:cloth}
	
\end{figure}

\subsection{Application to Multi-attribute Classification}
\label{sect:apptoatts}

The proposed NA model which is implemented as a linear layer can be flexibly plugged into any standard network architecture, and the proposed recursive learning strategy can be adopted for general network training. To show this, we extend our proposed approach to multi-attribute learning, i.e., cloth attributes classification, which is more challenging than a single attribute classification task,  especially in the noisy scenario. Fig.~\ref{fig:cloth} shows some examples from the cloth dataset \cite{huang2015cross}, there are five attributes for each sample~(Clothes Category, Color, Collar Shape, Sleeve Length, and Button). As some attributes, such as Color and Sleeve Length, are easy to be recognized in most cases and others, such as the type of Button, can be quite puzzled, the noise levels of the five attributes may vary a lot. Instead of training a whole CNN model for each attribute as in \cite{abdulnabi2015multi}, we append multiple NA models for various attributes on top of the shared base network. Let $\{x_n,\bm{y}_n\}$ be a sample with $K$ attributes, i.e., $\bm{y}_n\in\mathbb{R}^{K\times1}$, and the $k$-th attribute has $C_k$ classes, i.e., $\bm{y}_{n,k}\in\{1, 2, \dots, C_k\}$. Then for each attribute we employ $M$ noise-specific units parameterized by $Q\in \mathbb{R}^{C_k\times C_k}$. The recursive learning procedure is adopted similarly to that of the single attribute classification, in which the training supervisions are updated attribute-wisely. The network cost function is calculated independently for each attribute, for example, the negative log-likelihood loss of the $c$-th attribute in the first iteration/stage is: 
\begin{equation}
L_k(\theta,Q^k)=-\frac{1}{N}\sum_{n=1}^{N}\log p(y=\tilde{\bm{y}}_{n,k}|x_n, \theta, Q_{max}^k)
\end{equation}

The cost function for the network with combined training supervisions is similar to the one defined in single attribute task. The only difference is that for the multi-attribute task, there are $K$ classifiers taking as input the shared CNN features, each of which is followed by a NA model. The back-propagation follows the same pipeline with the single attribute task.        

\section{Experiments}
\label{sect:experiment}

To evaluate the Noise-Attention model and the recursive learning strategy, we conduct experiments on two popular datasets: CIFAR-10~\cite{krizhevsky2010convolutional} and SVHN~\cite{netzer2011reading} for single attribute classification and a real-world clothes dataset~\cite{huang2015cross} for multiple attributes classification. Since the first two datasets are noise free, we randomly select and flip the labels of some training samples. Let $\rho$ denote the noise level of the training set, i.e., the percentage of mislabeled images in the training set. For the real-world dataset, since the images are collected from websites with user-supplied tags, it naturally contains a certain level of mislabeled samples. 

\textbf{Baselines:} Several recently proposed noisy label learning methods are chosen as our baselines: 

\begin{enumerate}
	\item Bottom-up~\cite{sukhbaatar2014learning}: A CNN model contains a single noisy layer on the top to learn noisy distribution.
	\item Joint CNNs \cite{Xiao_2015_CVPR}: A combined CNN model consists of two branches.
	\item Dropout-Reg \cite{jindal2016learning}: A regularization scheme is proposed and applied to the output of CNN to prevent it from learning the noisy labels directly.
	\item SEC-CNN \cite{liu2017self}: A confidence policy is introduced to correct the wrong label by the max-activated output neuron of the CNN.
	\item Iterative \cite{wang2018iterative}: A Siamese network is designed to encourage clean labels and noisy labels to be dissimilar, which is learned in an iterative fashion.
	\item GeneralizedCCE \cite{zhang2018generalized}: A grounded set of noise-robust loss functions is introduced, they can be applied to any CNN architectures and algorithms.
	\item ResistanceNN \cite{drory2018resistance}: Inspired by k-NN classifier, Drory \textit{et al.} propose to predict the label of a test sample based on a neighborhood of the training samples, which is applied to the problem of noise label classification. 
	\item BundleNet \cite{li2018bundlenet}: A sample-bundle module is designed to utilize sample correlations by constructing bundles of samples class-by-class, it can be applied to any CNNs by replacing the conventional input layer with it.
	\item Distillation \cite{li2017learning}: A unified distillation framework is designed to relieve the influence of label noise by ``side'' information, including a small clean dataset and label relations in the knowledge graph.
	\item Base model: the vanilla CNN.
\end{enumerate}        

The proposed attention network and the recursive learning strategy are implemented on Caffe~\cite{jia2014caffe}. We resort to Caffe's CIFAR-full model as the base network for the first two datasets, and DARN \cite{huang2015cross} for real-world cloth dataset. The attention network containing NA model is denoted as ``NAtt.'', and it is denoted as ``NAtt. $+$ Rec.'' if it is trained using recursive learning, and as ``NAtt. $+$ Iter.'' if it is trained using conventional training process.

\begin{figure}[t]
	\centering
	\subfigure{\includegraphics[width=0.45\linewidth]{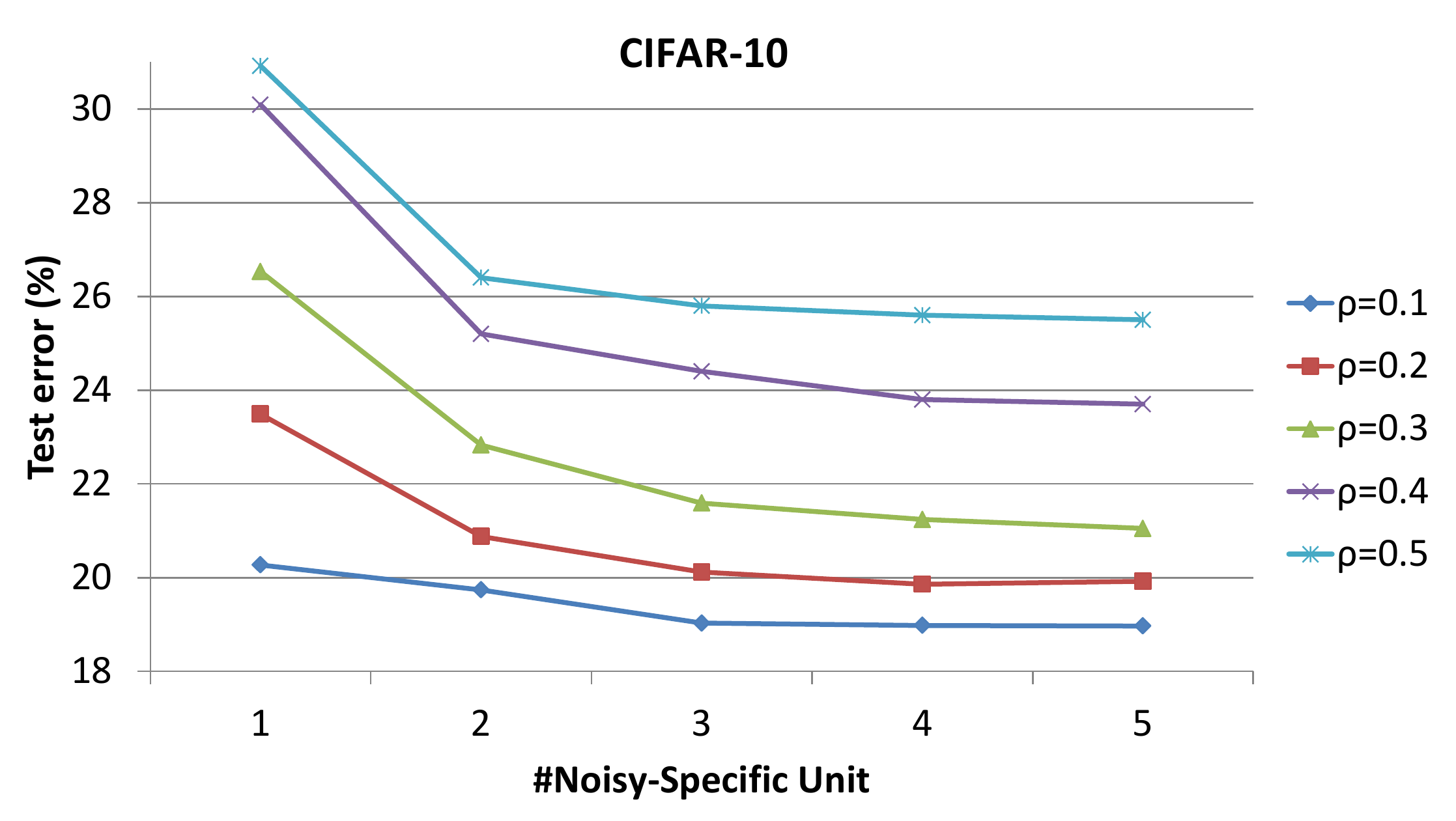}\label{fig:cifar01}}
	\subfigure{\includegraphics[width=0.45\linewidth]{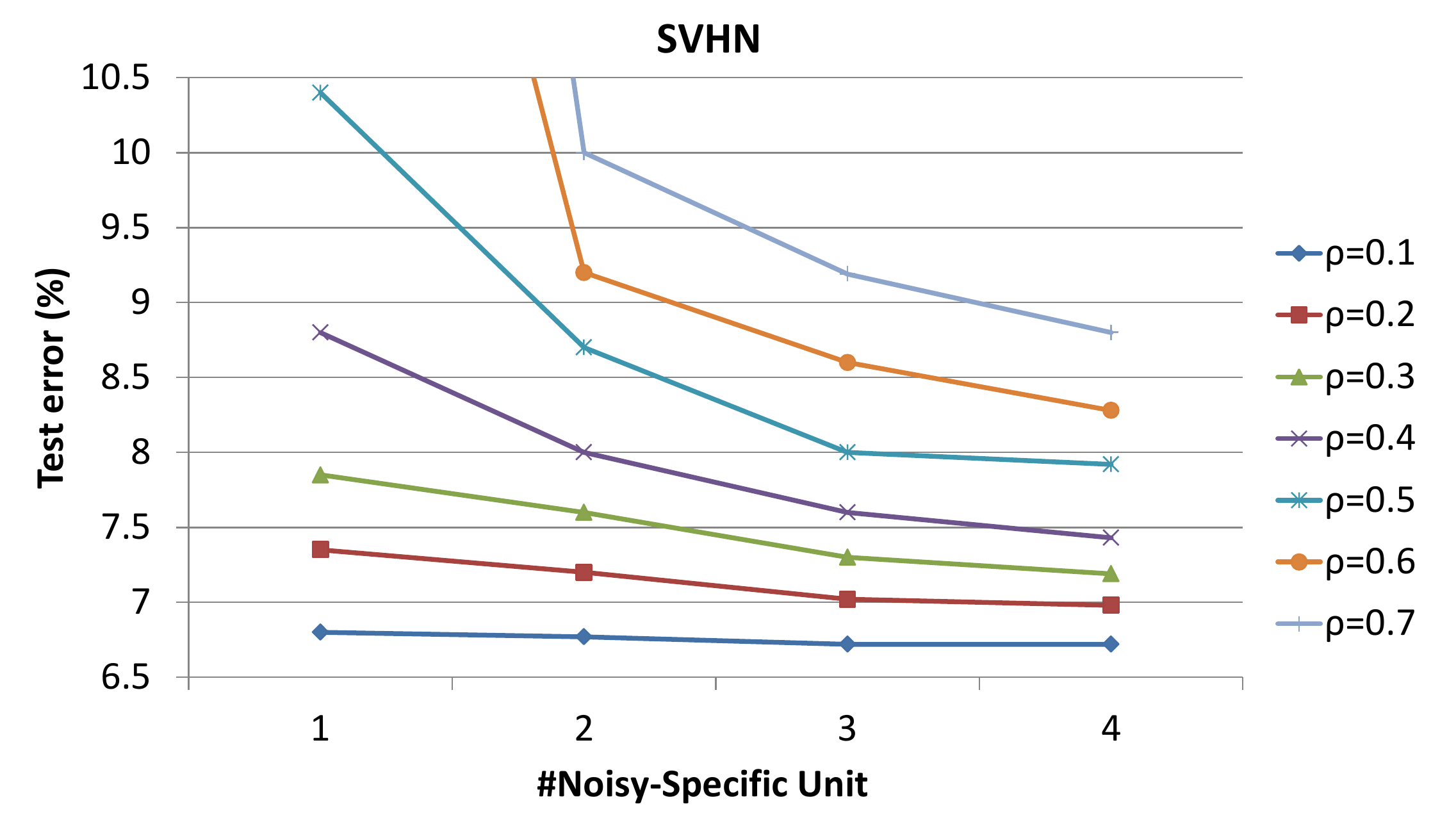}\label{fig:svhn01}} 

	\caption{Test errors (\%) on CIFAR-10 and SVHN in terms of number of Noise-Specific Units in NA model.}\label{fig:number-ns}
\end{figure}

\subsection{Manually Flipped Noise}
\label{sect:flippednoise}

\textbf{Dataset Details:} CIFAR-10 contains 60k $32\times32$ color images from 10 classes, 50k for training and 10k for testing. SVHN contains over 600k images of house number digits, 100k of which are used for training and 26k for testing. It has 10 classes, one for each digit. We randomly select and flip training samples with noisy level $\rho=[0.1, 0.2, 0.3, 0.4, 0.5, 0.6, 0.7]$.

\textbf{Influence of number of noise-specific units.} The number of noise-specific units contained in NA model is adjustable according to the dataset and noisy level. In this part, we will illustrate the influences of different numbers of units. As stated in Sect. \ref{sect:noisemodel}, a noise-specific unit will be added to the NA model if the training loss converges based on current units. Before that, the base model of CIFAR-10 is trained for 30 epochs, and that of SVHN for 20 epochs so that the noisy information is fully learned by the base networks. Then we keep training networks while adding noise-specific units to the NA model one by one until the performance improvement is limited.      

Fig.~\ref{fig:number-ns} shows the test errors of the proposed attention network with different numbers of noisy unit $\{Q_m\}_{m=1}^M$. Recall that in our definition $Q_1$ is fixed as $I$, so the attention network will shrink into the base model if it contains only one noise-specific unit. From the two figures, one can see that adding more noise-specific units has slight improvement in performance when noise level is low, such as $\rho=0.1$ on CIFAR-10 and $\rho=0.1,0.2$ on SVHN. This is reasonable since CNN models intuitively have strong learning ability to handle a small portion of noisy labels. However, the noisy tolerance capacity of the base model decreases sharply with the increasing number of noisy samples. Although our proposed noise-attention network is also degraded by high-level noise, the superiority over the base model is remarkable, e.g., test errors reducing $6\%$ on CIFAR-10 at $\rho=0.4$. Theoretically, the more noise-specific units are added, the better of the classification results will be. However, the improvement slows down, so we use $M=5$ noise-specific units for CIFAR-10 and $M=4$ for SVHN in the following experiments to evaluate the recursive learning strategy.  

\begin{figure}[t]
	\centering   
	\subfigure{\includegraphics[width=0.45\linewidth]{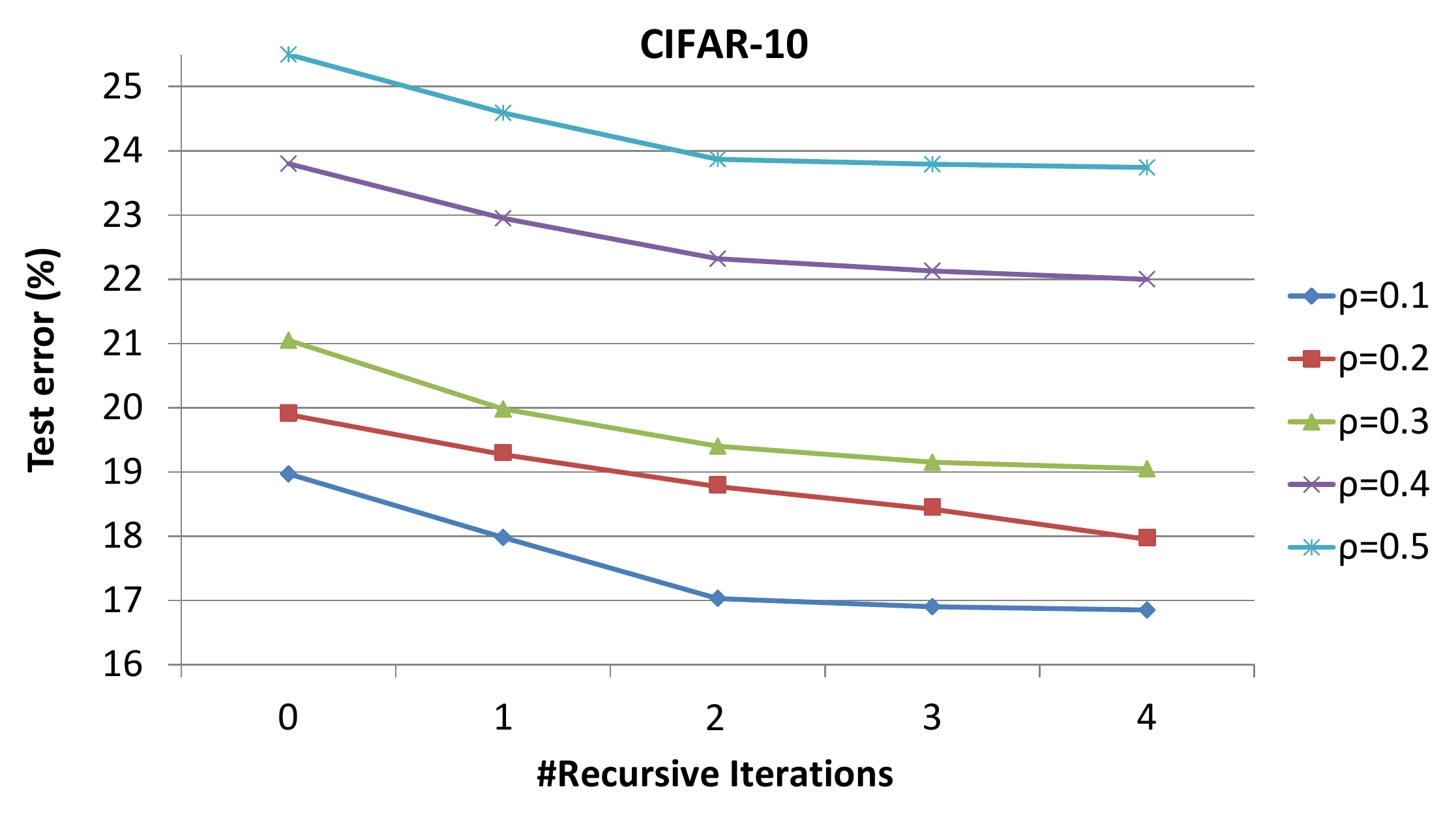}\label{fig:cifar02}}
	\subfigure{\includegraphics[width=0.45\linewidth]{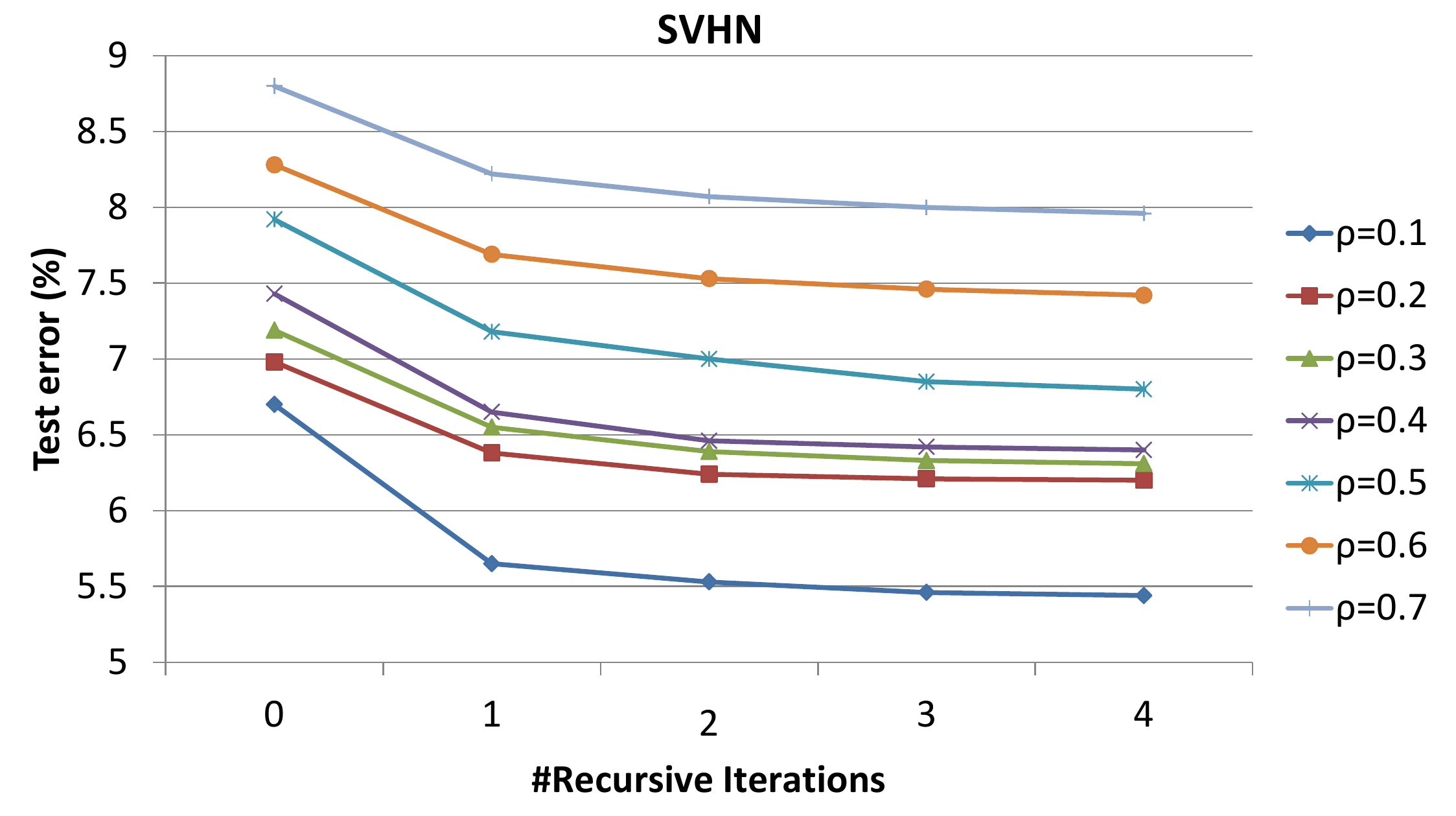}\label{fig:svhn02}} 
  
	\caption{Test errors (\%) on CIFAR-10 and SVHN in terms of number of Recursive Iterations.}\label{fig:number-iteration}
\end{figure} 

\textbf{Influence of number of recursive iterations.} Although the attention network has achieved satisfactory performance, it can be further boosted by the recursive learning strategy. The outputs of the well-trained attention network with multiple noise-specific units will be combined with the given labels as the new training supervisions for recursive learning process. The number of noise-specific units is fixed as $M=5$ for CIFAR-10 and $M=4$ for SVHN. The coefficient $\alpha$ for the given labels in $t$-th iteration is set as $\alpha=0.8^t$. Instead of emphasizing adjusting and learning NA model as in the first stage, we focus on the whole network learning in the recursive learning stage.    

Fig.~\ref{fig:number-iteration} shows the test errors changing with the number of recursive iterations. The results of the network in the first stage are denoted by $t=0$. The test errors decrease dramatically after one iteration, and the trend slows down with more iterations. For CIFAR-10, the proposed recursive learning strategy has an average improvement with about $2\%$ after training four iterations. For SVHN, after four iterations, the model trained at $50\%$ noisy level even beats the model with only $10\%$ mislabeled samples present in the training set.

\begin{figure}
	\centering
	\subfigure[CIFAR-10 at $\rho=0.3$.]
	   {\includegraphics[width=0.18\linewidth]{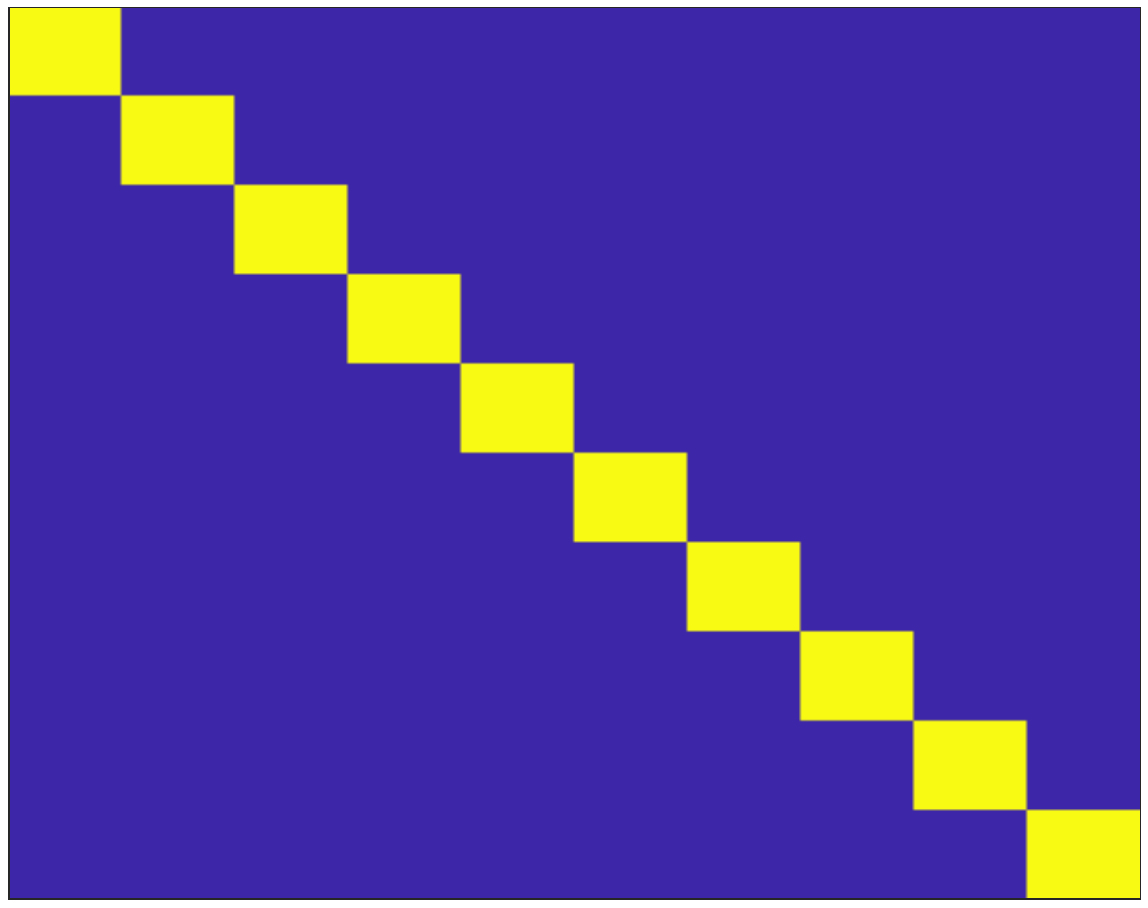}
	   \includegraphics[width=0.18\linewidth]{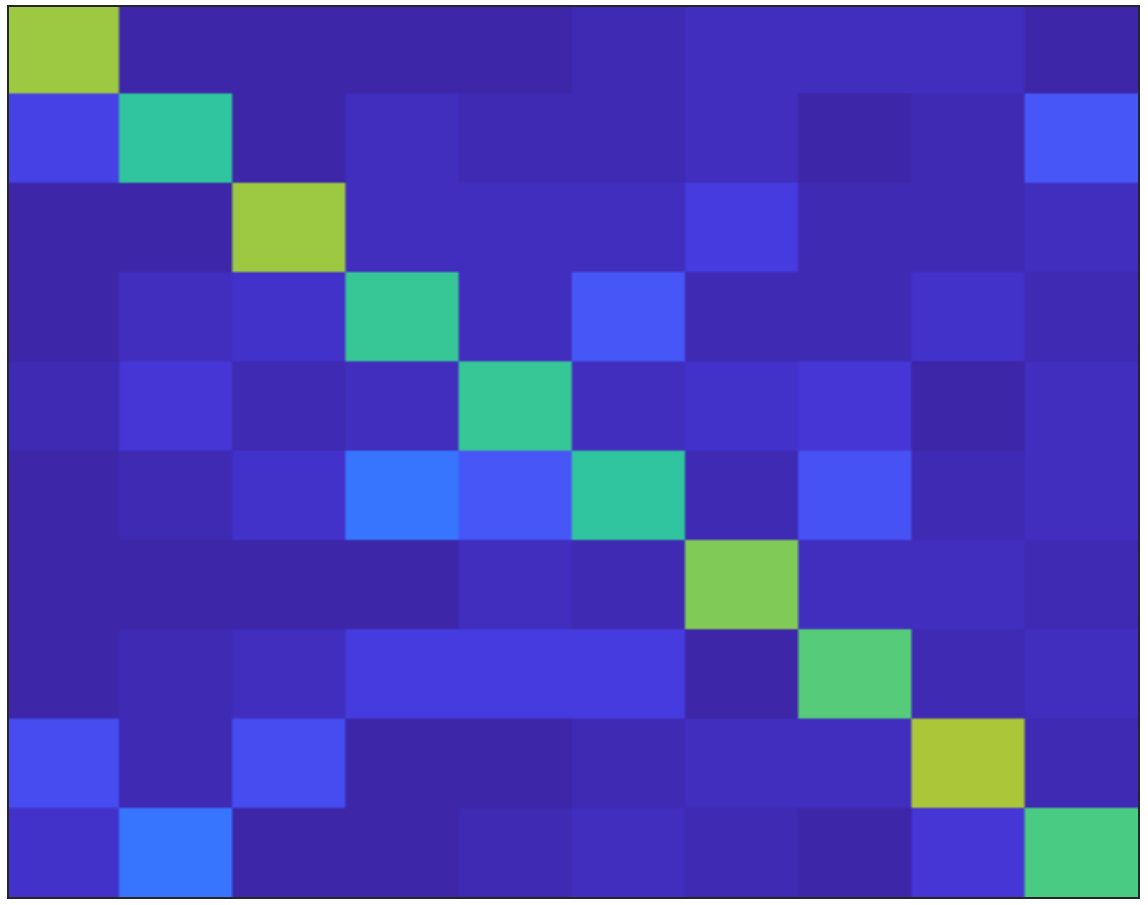}
	   \includegraphics[width=0.18\linewidth]{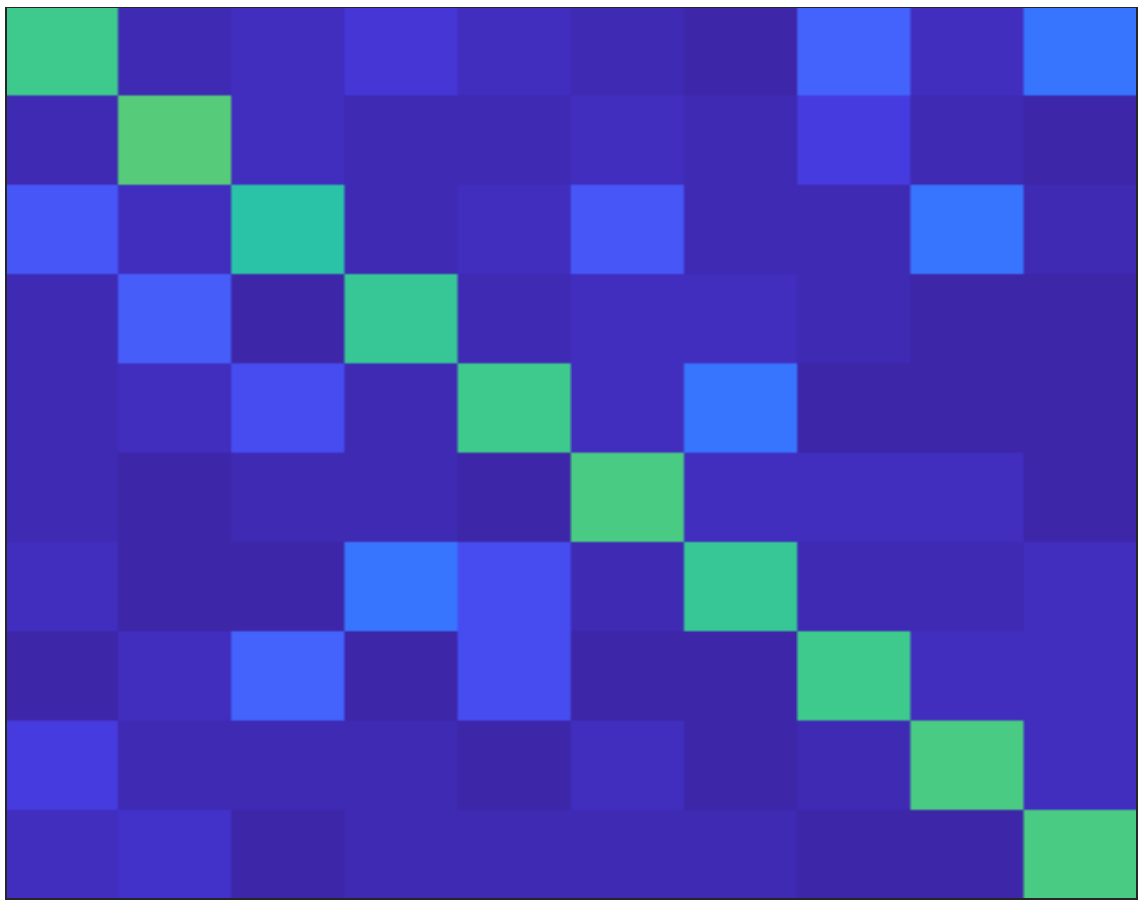}
	   \includegraphics[width=0.18\linewidth]{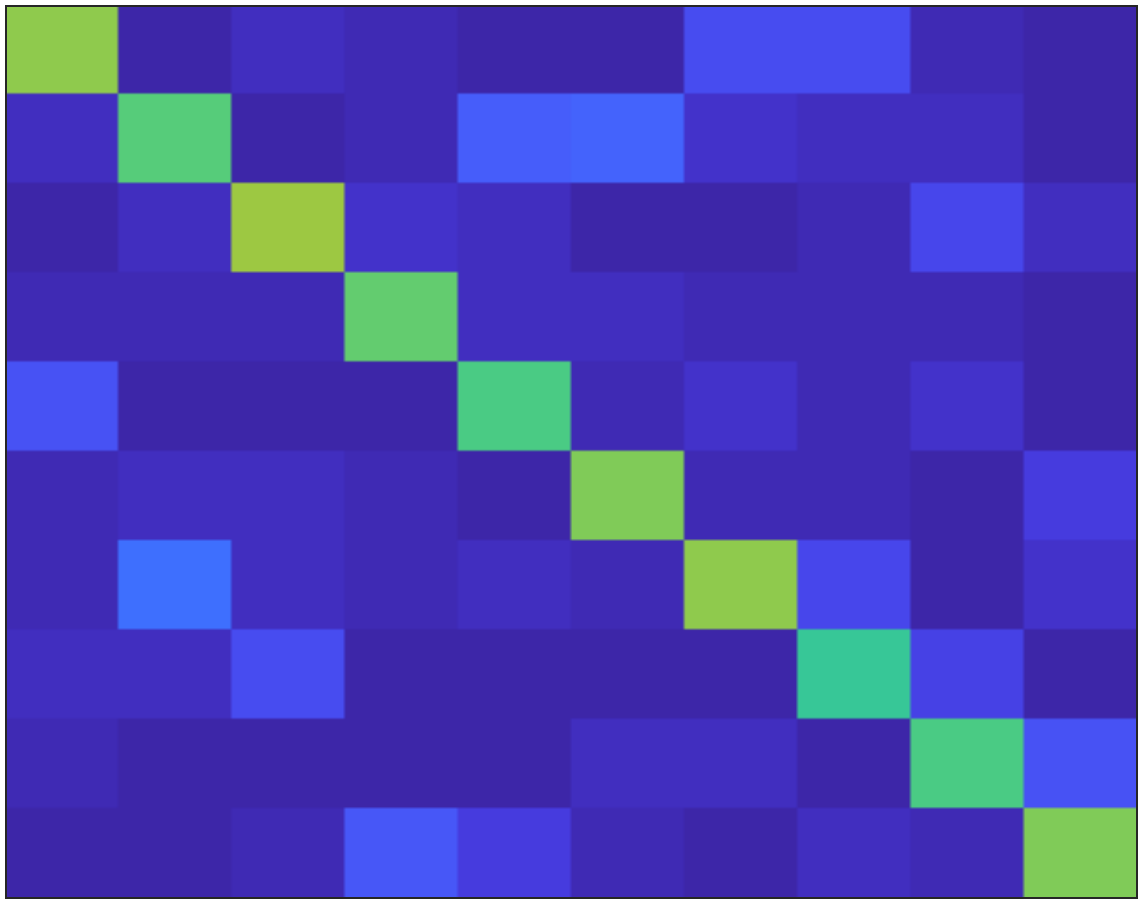}
	   \includegraphics[width=0.18\linewidth]{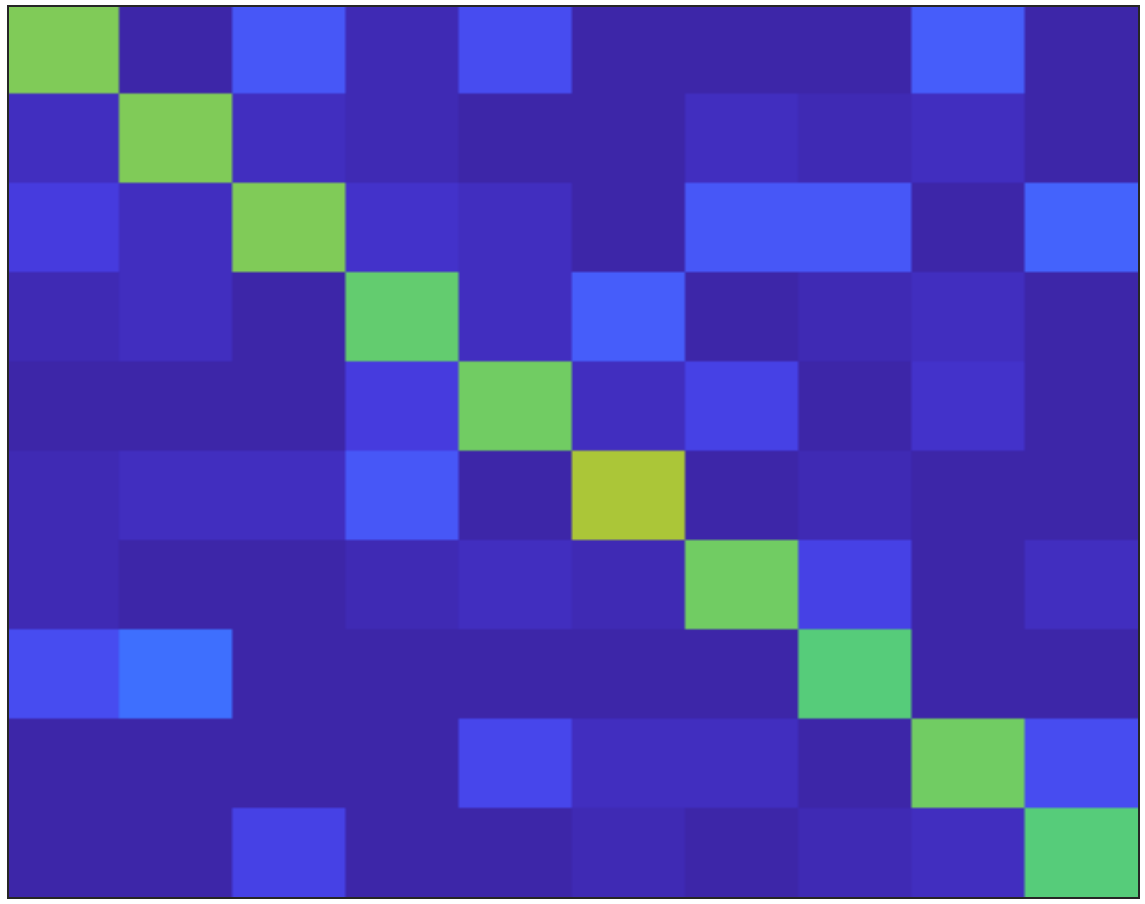}} \\
	
	\subfigure[SVHN at $\rho=0.5$.]
	   {\includegraphics[width=0.18\linewidth]{cq1.pdf}
	   \includegraphics[width=0.18\linewidth]{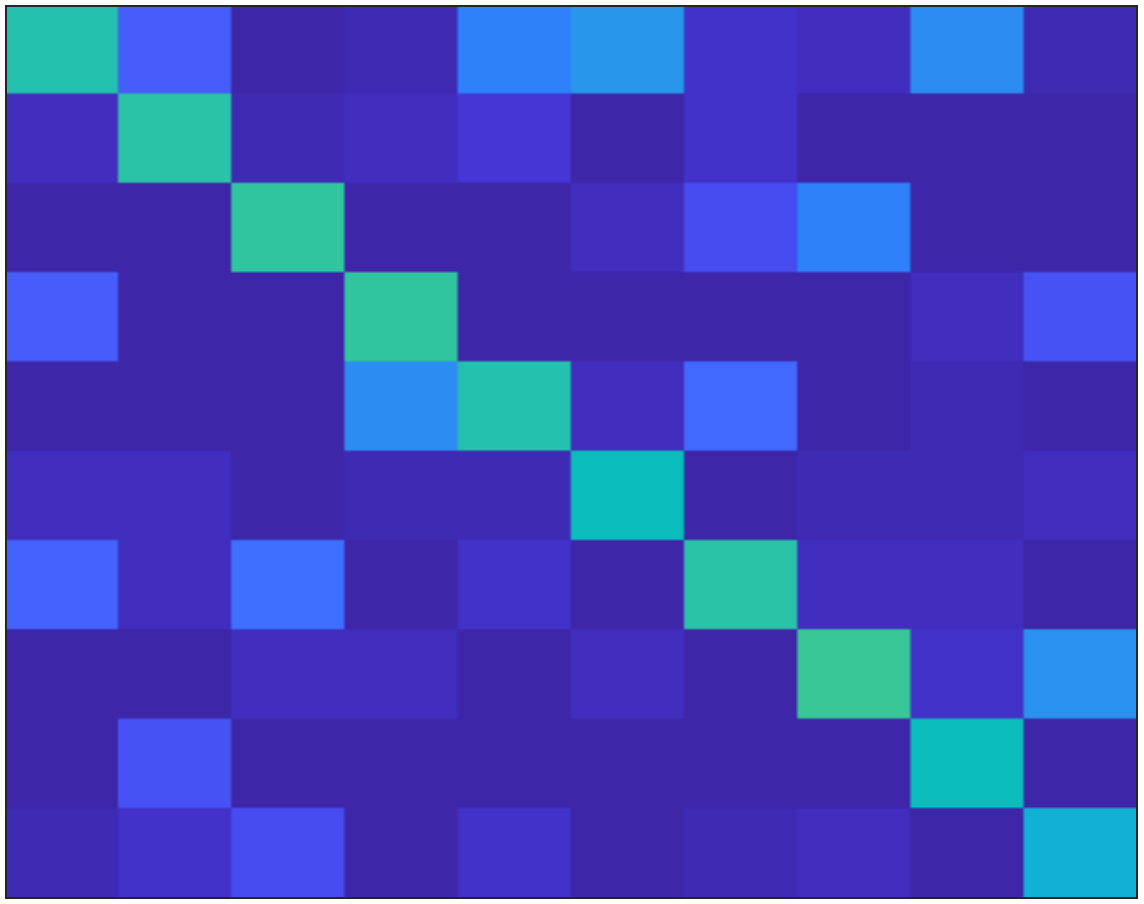}
	   \includegraphics[width=0.18\linewidth]{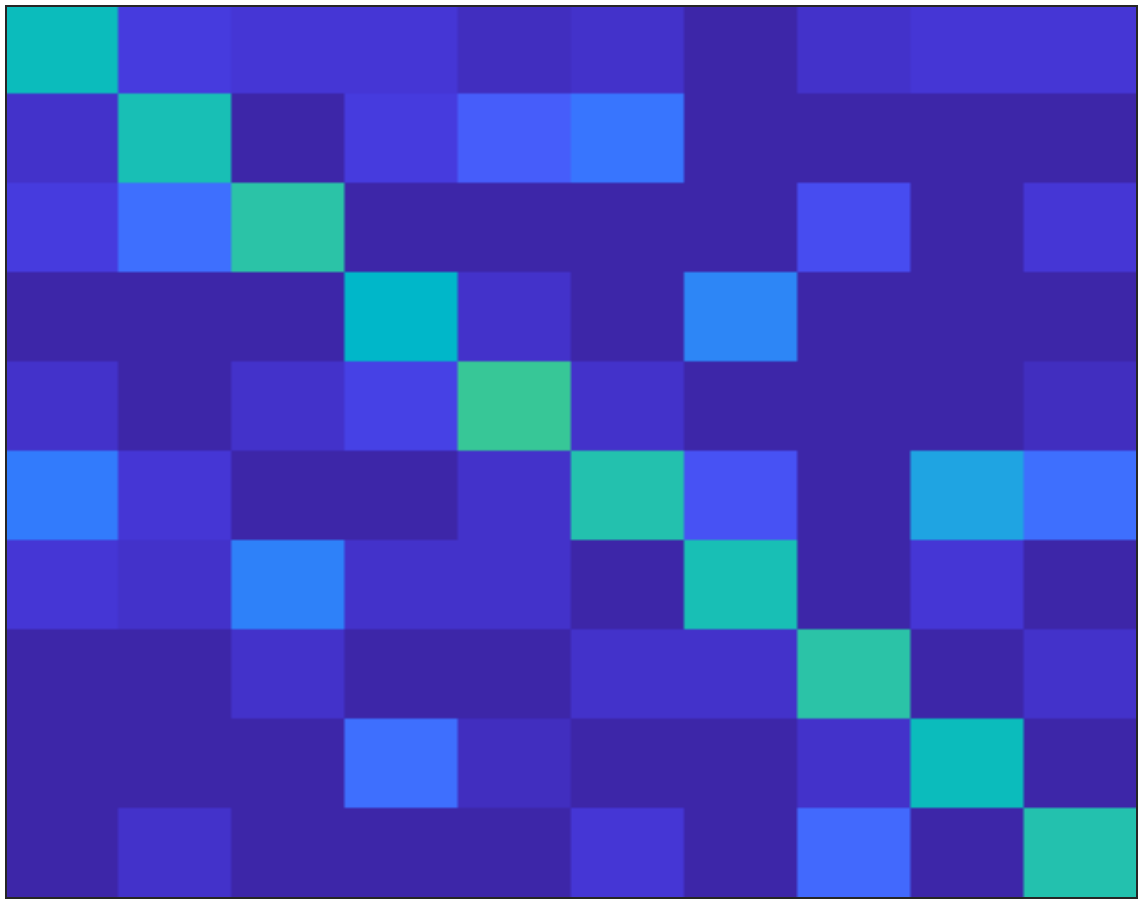}
	   \includegraphics[width=0.18\linewidth]{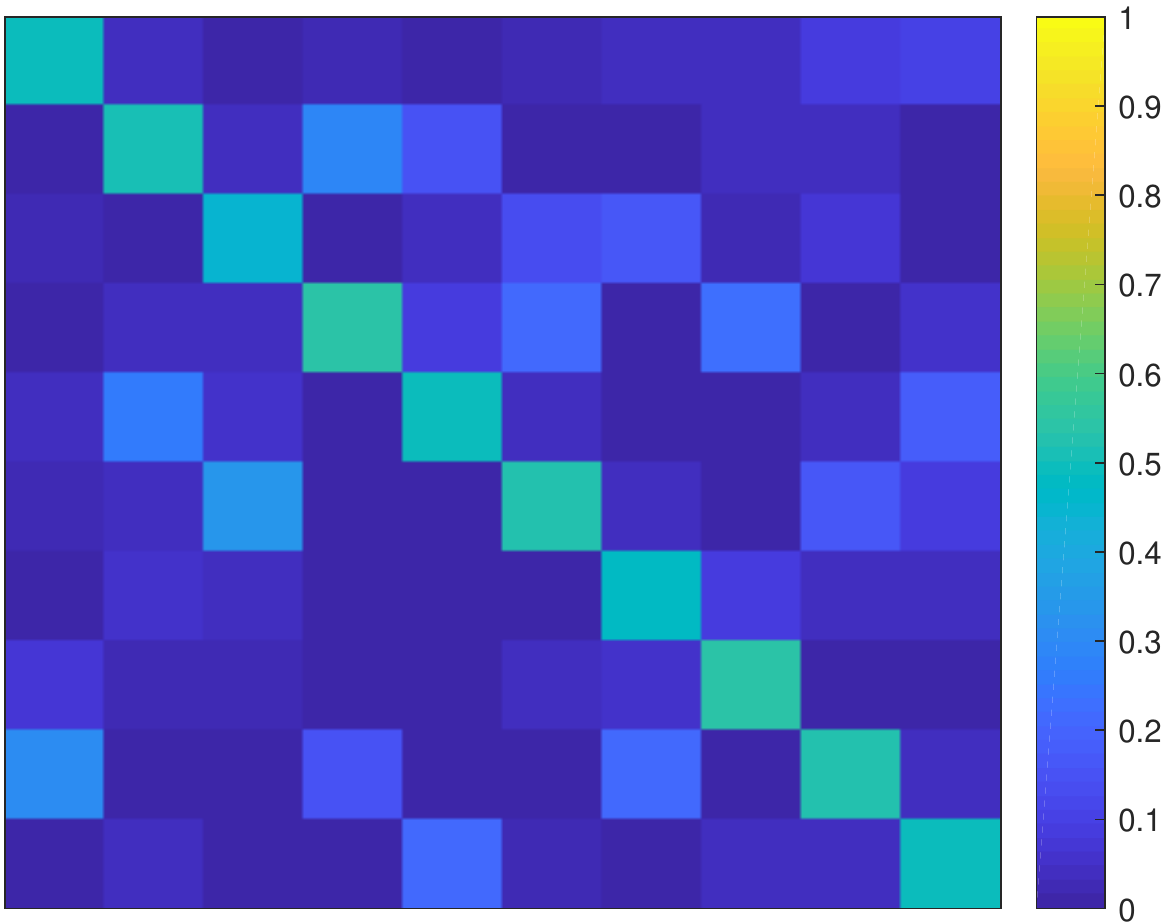}}   
	
	\caption{Visualization of noise-specific units. The first column represents $Q_1=I$, the rest ones show the learned noise-specific units.}
	\label{fig:visulization}
	
\end{figure}

\textbf{Visualization of noise-specific units.} We visualize the learned noise-specific units in Fig. \ref{fig:visulization} for CIFAR10 at $\rho=0.3$ and SVHN at $\rho=0.5$. Figures in the first column represent $Q_1$ which is fixed as an identity matrix. The rest columns show the learned noise-specific units, with the diagonals indicating the correctly labeled ratios. Due to the gradually learning strategy, the learned noisy distributions are diverse to each other so that different noisy information in training samples are able to be better modeled.

\begin{table}[t]
	\centering
	\caption{Test errors in the presence of noisy labels.}
	
	\subtable[CIFAR-10.]{
		\resizebox{7cm}{!}{
			\begin{tabular}{c|ccccc}
				\hline
				Noise Level & 10\% & 20\% & 30\% & 40\% & 50\% \\
				\hline
				Base model  & 20.27   & 23.49 & 26.53 & 30.09 & 30.92 \\
				Bottom-up \cite{sukhbaatar2014learning} & 19.74 & 20.88 & 22.83 & 25.2 & 26.4 \\
				Joint CNNs \cite{Xiao_2015_CVPR} & 19.6 & 20.5  & 22.8 & 25.0 & 26.47 \\ 
				Dropout-Reg \cite{jindal2016learning} & - & - & 25.4 & - & 31.28 \\
				SEC-CNN \cite{liu2017self} & 21.99 & 26.75 & 30.09 & 38.35 & 44.72 \\
				Iterative \cite{wang2018iterative} & - & 18.64 & - & 21.85 & -\\
				{GeneralizedCCE \cite{zhang2018generalized} } & {18.39} & {19.12} & {19.89} & {21.88} & {-} \\
				{ResistanceNN \cite{drory2018resistance} } & {18.7} & {19.2} & {20.0} & {26.5} & {56.0} \\
				{BundleNet \cite{li2018bundlenet} } & {11.78} & {-} & {19.66} & {-} & {43.26}\\
				{Distillation \cite{li2017learning} } & {15.24} & {17.60} & {19.51} & {21.89} & {24.16} \\
				\hline
				NAtt.    & 18.97 & 19.92 & 21.05 & 23.7 & 25.5 \\
				NAtt. $+$ Iter. & 18.27 & 19.95& 21.1&23.28 & 25.1 \\
				NAtt. $+$ Rec. & 16.95 & 17.97 & 19.13 & 22.02 &  23.83\\
				\hline
			\end{tabular}
		}
	}\\
	
	\subtable[SVHN.]{
		\resizebox{8cm}{!}{  
			\begin{tabular}{c|ccccccc}
				\hline
				Noise Level  & 10\% & 20\% & 30\% & 40\% & 50\% & 60\% & 70\% \\
				\hline
				Base model & 6.8  & 7.35 & 7.85 & 8.8  & 10.4 & 16.0 & 22.0 \\
				Bottom-up \cite{sukhbaatar2014learning} & 6.77 & 7.2 & 7.6 & 8.0 & 8.7 & 9.2 & 10.0 \\
				Joint CNNs \cite{Xiao_2015_CVPR} & 6.75 & 7.08 & 7.41 & 7.67 & 8.7 & 9.21 & 9.79 \\ 
				Dropout-Reg \cite{jindal2016learning} & 7.04 & 7.28 & 7.91&8.73 &10.3 &15.27 & 18.6\\
				{Distillation \cite{li2017learning} } & {5.54} & {6.21} & {6.27} & {6.53} & {6.71} & {7.5}&{8.01}\\
				\hline
				NAtt.         & 6.7 & 6.98 & 7.19 & 7.43 & 7.92 & 8.28 & 8.8 \\
				NAtt. $+$ Iter. & 6.83 & 7.1 & 7.21 & 7.37 & 7.76 & 8.14 & 8.35 \\
				NAtt. $+$ Rec. & 5.44& 6.2 & 6.31& 6.4& 6.8& 7.42& 7.96 \\
				\hline
			\end{tabular}
		}
	}
	\label{tab:cifar-svhn}
\end{table}

\textbf{Comparison with state-of-the-art.} To fairly evaluate our proposed approach, we compare the results with related literature in Table~\ref{tab:cifar-svhn}. Our results listed are achieved with $M=5$ noisy units for CIFAR-10 dataset and $M=4$ for SVHN dataset, $t=4$ recursive iterations for both datasets. To validate the effectiveness of the proposed recursive learning method, we also list the results achieved by an attention network learned using traditional training process but with the same epochs as in ``NAtt. $+$ Rec.', which we denote as ``NAtt. $+$ Iter.''. 

{Table~\ref{tab:cifar-svhn}(a) lists the test errors on CIFAR-10 with considered methods at different label noise levels. All results are reported in the corresponding publications, except Distillation \cite{li2017learning} which is implemented based on the released code. Our proposed noise-attention network (``NAtt'') outperforms the majority of baselines except the Iterative \cite{wang2018iterative}. After four-rounds training using recursive learning strategy (``NAtt+Rec.''), our proposed method beats Iterative \cite{wang2018iterative} by about $1.4\%$ at $\rho=20\%$ and about $0.8\%$ at $\rho=40\%$. The performance of GeneralizedCCE \cite{zhang2018generalized}, ResistanceNN \cite{drory2018resistance}, and Distillation \cite{li2017learning} is on par with our method at the low level of label noise, while these methods are inferior to ours under the high proportion of label noise. Note that the Distillation \cite{li2017learning} even uses extra clean samples during training, the size of the clean dataset is one-quarter of the training set on both Cifar10 and SVHN.} The results achieved by ``NAtt. $+$ Iter.'' are similar to these of ``NAtt.'' as the network in ``NAtt. $+$ Iter.'' is exactly the one in ``NAtt.'' except being trained with more epochs, however more training epochs does not always bring better performance. Although sharing the same architecture and running with equal epochs, the attention model trained by ``NAtt+Rec.'' performs favorably against that by ``NAtt. $+$ Iter.'' with averagely $2\%$ improvement, which demonstrate the effectiveness of our proposed recursive leaning method.

Table~\ref{tab:cifar-svhn}(b) lists the results on SVHN dataset, among which the Joint CNNs \cite{Xiao_2015_CVPR}, Dropout-Reg \cite{jindal2016learning} and Distillation \cite{li2017learning} are implemented by ourselves following their released code. With extra clean data, Distillation \cite{li2017learning} performs similarly to our method. The performance obtained by Joint CNNs and Bottom-up is similar on this dataset, while is superior over Dropout-Reg \cite{jindal2016learning} and inferior to the proposed attention network (``NAtt.''), especially at high noise levels. More training epochs on attention network (``NAtt. $+$ Iter.'') even degrades the model at low-level noise due to network overfitting. For both datasets, the base model aggravates severely with the increasing number of noisy samples, while our noise-attention network is more stable and can be further improved with the recursive learning method. 

\begin{table*}[t]
	\centering
	\caption{Test errors~(\%) of the considered methods on clothes images.}
	
	\resizebox{14cm}{!}{  
		\begin{tabular}{c|cccccc}
			\hline
			SVHN & Clothes Category & Clothes Color & Collar Shape & Sleeve Length & Clothes Button & ALL.\\
			\hline
			Base model  &  9.27 & 30.19 & 33.02 & 17.56 & 31.14 & 40.39  \\
			Bottom-up \cite{sukhbaatar2014learning} & 8.5 & 26.78 & 32.2 & 15.1 & 30.73 & 39.17 \\
			Joint CNNs \cite{Xiao_2015_CVPR} & 6.93 & 23.88 & 29.7 & 12.04 & 27.7 & 36.46 \\
			Dropout-Reg \cite{jindal2016learning} & 8.98 & 27.26& 33.4& 15.81&30.2 & 38.47\\
			{Distillation \cite{li2017learning}} & {8.31} & {23.74} & {28.57} & {13.35} & {26.88} & {35.19}\\
			\hline
			NAtt.    & 7.1 & 24.8 & 30.38 & 12.85  & 28.6 & 37.29\\
			NAtt. $+$ Iter. &7.13 & 24.16 &30.44 & 12.7 & 27.96 & 36.14 \\
			NAtt. $+$ Rec. & 6.4 & 22.0 & 29.07 & 11.0 & 26.8 & 34.8 \\
			\hline
		\end{tabular}
	}
	
	\label{tab:cloth}
\end{table*}

\begin{figure}
	\centering
	\subfigure{\includegraphics[width=0.45\linewidth]{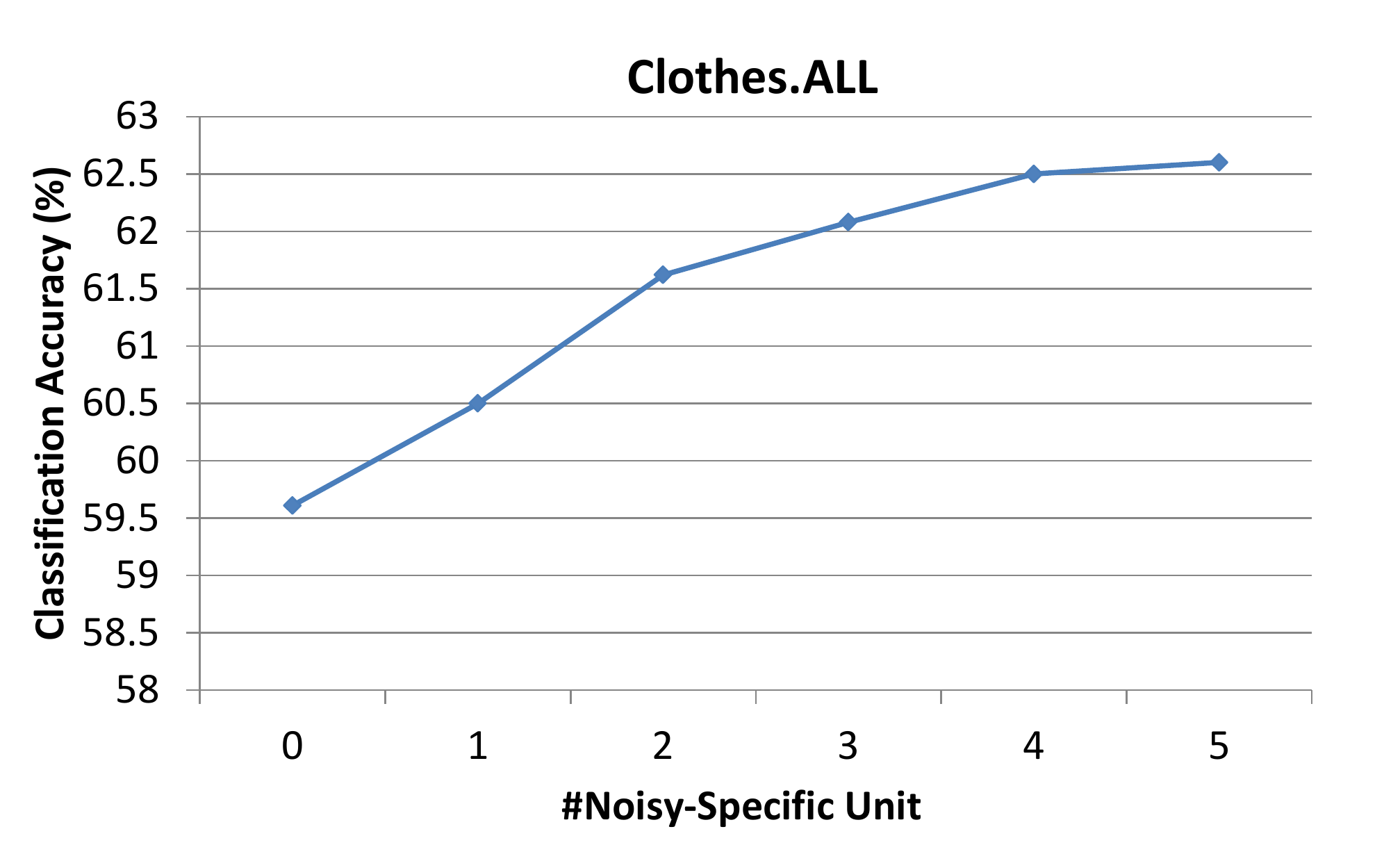}}
    \subfigure{\includegraphics[width=0.45\linewidth]{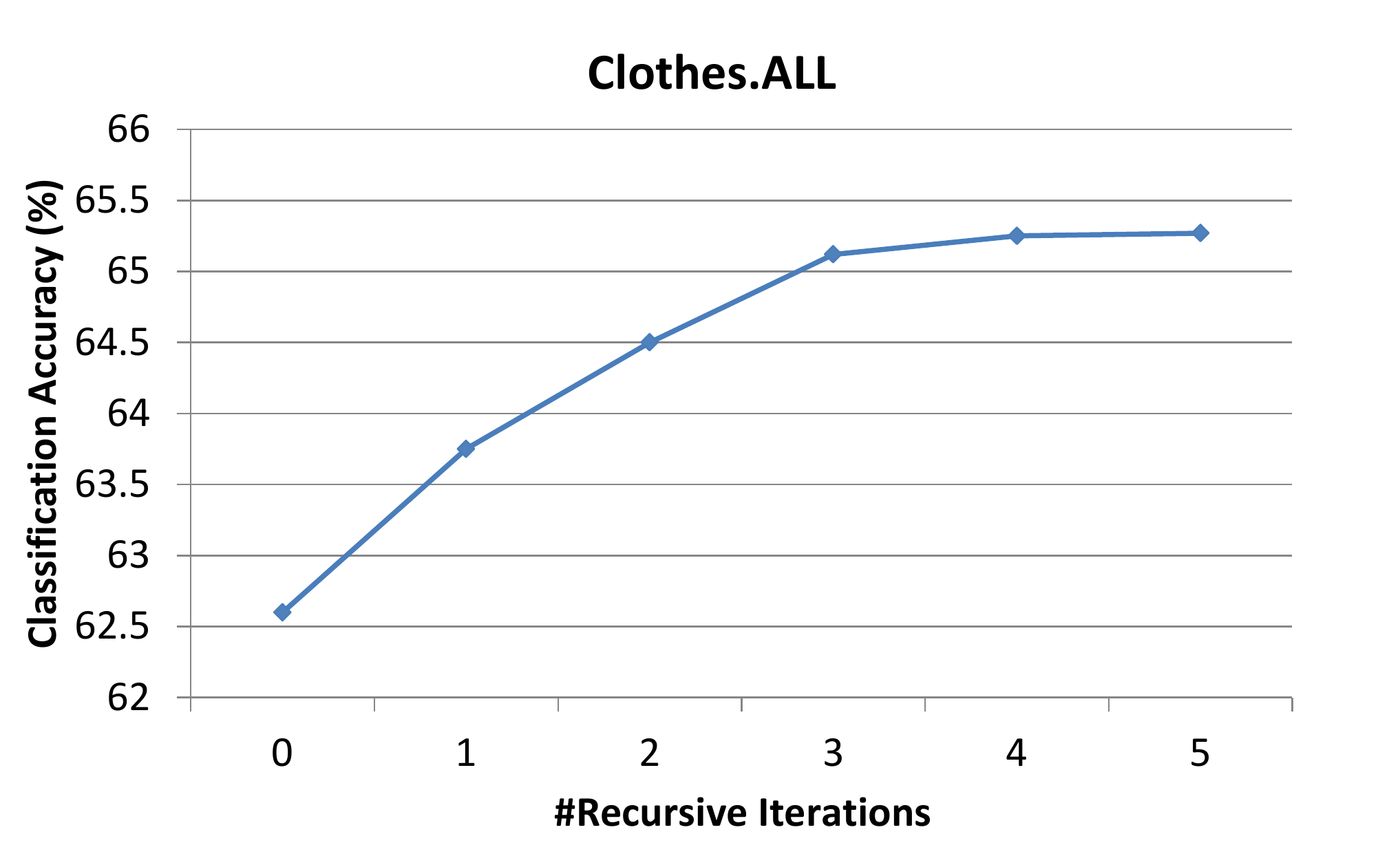}}	
	\caption{Test errors (\%) on real-world cloth dataset in terms of number of noise-specific units in Noise-Attention model and number of recursive iterations.   
	}\label{fig:expcloth}
	
\end{figure}

\subsection{Real-world Clothing Noise}
\label{sect:realnoise}
We further evaluate the considered methods on a real-world cloth dataset, which is an open set introduced in~\cite{huang2015cross}, the images of which are crawled from online shopping websites such as TMALL. In particular, each image has 9 semantic attribute categories, and there are more than a hundred attribute labels in total. We select 5 attribute categories in our experiment, the total number of attribute labels is $56$. Since some images are taken and labeled by professionals, some are user-supplied, it is highly possible that the noisy distributions of different attribute categories will vary a lot. We use 118,057 images for training, and another 2000 images which are manually purified by us for testing. The network from~\cite{huang2015cross} is chosen as the base model, which is trained for 30 epochs till convergence. Then the noise-specific units is gradually added to the NA model one by one until the performance improvement is limited. There are $6$ noise-specific units contained in each NA model before being fine-tuned using recursive learning method, which lasts for another $5$ recursive iterations before stopping training. 

Fig.~\ref{fig:expcloth} shows the test errors~(\%) changing with the different numbers of noise-specific unit $\{Q_m\}_{m=1}^M$ and at different iterations of the recursive learning strategy. The ``.ALL'' in title means that the samples are true positive only if the $5$ attribute categories are all correctly predicted. As the noisy distributions of real-world images are rather complicated, the proposed attention network enjoys significant improvement over the base model by about $4\%$ with $M=6$. And this performance can be further enhanced by our recursive learning strategy. Fig.~\ref{fig:expcloth}(b) depicts the decreasing tendency of the test errors with the increasing number of recursive iterations. After only one round, the performance can be improved by more than $1\%$. We stopped training at $t=5$ for the improvement over the previous iteration is trivial. Compared with the Bottom-up model \cite{sukhbaatar2014learning}, which only sets a single noisy layer, the proposed noise attention network achieves $2\%$ improvement and after being fine-tuned using recursive learning strategy, the improvement rises to $4\%$. 

Table~\ref{tab:cloth} lists the text errors~(\%) achieved by considered methods on the individual attribute category, and ``ALL'' means that the five attribute categories of a sample are all predicted correctly. The results of Bottom-up \cite{sukhbaatar2014learning}, Joint CNNs \cite{Xiao_2015_CVPR}, Dropout-Reg \cite{jindal2016learning} and Distillation \cite{li2017learning} are implemented by us based on their released codes. From the table, we can see the performance achieved on different attribute categories varies a lot even using the same method. The possible reason is that some attribute categories are easy to be labeled, some may be sensitive to the environment. Among the considered methods, only the Joint CNNs~\cite{Xiao_2015_CVPR} performs slightly better than our attention network, but is far behind the network trained using our recursive learning strategy. Note that \cite{Xiao_2015_CVPR} is trained using an extra number of clean data, \cite{sukhbaatar2014learning}, \cite{jindal2016learning} and our attention network however are all trained using noisy data. { Although the Distillation \cite{li2017learning} performs on par with our ``NAtt. $+$ Rec.'', it requires extra clean data for training, an extra set of 2000 images with correct labels is used for pretraining the Distillation model.} The proposed method and Joint CNNs outperform Bottom-up by a large margin as it favors uniformly distributed noise. Similar to the results in manually flipped label noise scenario, the model trained in ``NAtt. $+$ Iter.'' performs similarly to ``NAtt.'', while is significantly inferior to ``NAtt. $+$ Rec.''. Although the three baselines and the base model are more efficient during training, the testing phase of the proposed attention network trained with recursive learning strategy is similar as traditional classification network with only the base network being used to evaluate.

\section{Conclusion}
\label{sect:conclusion}
In this paper, we propose an attention-aware noisy label learning approach to address the problem of image classification in the presence of label noise. Firstly, a Noise-Attention model is designed to learn the diverse noisy label information in the training set. It consists of multiple noise-specific units, each of which pays attention to a specific noisy label distribution. To strengthen the learning ability of the attention network, we further introduce a recursive learning strategy which could integrate the high-level knowledge from another network without introducing more parameters. In the learning process, a well-trained attention network is able to bootstrap itself by the combined training supervisions of given labels and outputs from the previous iteration. To validate the proposed methods, we conduct experiments on both synthesized label noise datasets and real-world cloth dataset with multiple attributes. The ablation studies on various numbers of noise-specific units and numbers of recursive iterations show the effectiveness of the Noise-Attention model and recursive learning strategy. The extensive experiments on images with deliberately flipped label noise and real-world label noise show the generalization of the proposed approach.

\bibliographystyle{ieee_fullname}
\bibliography{DRLHashing}
\end{document}